%% file: main.tex
\documentclass[10pt,conference]{resources/IEEEtran}
\usepackage[T1]{fontenc}
\usepackage[utf8]{inputenc}
\usepackage[english]{babel}
\AtBeginDocument{%
\providecommand\BibTeX{{%
\normalfont B\kern-0.5em{\scshape i\kern-0.25em b}\kern-0.8em\TeX}}}

\usepackage{hyperref}
\usepackage{cite}

\usepackage{amsmath,mathtools,wrapfig,amssymb,amsfonts} %amsmath already included
\usepackage[noend]{algpseudocode}
\usepackage{algorithmicx, algorithm}
\usepackage{graphicx} %graphicx already included
\usepackage{textcomp}
\usepackage{xcolor}
\usepackage{booktabs}

\usepackage{tikz}
\usepackage{pgfplots}
\pgfplotsset{width=7cm,compat=1.17}
\usetikzlibrary{arrows,decorations.markings, positioning, fit, calc}
\usepackage{subcaption}
\usepackage{tabu}
\usepackage{enumitem}

\usepackage{resources/mathcommands}
\usepackage{resources/scheduling}

\usepackage{amsthm}

\theoremstyle{definition}

\theoremstyle{remark}

\algnewcommand\algorithmicforeach{\textbf{for each}}
\algdef{S}[FOR]{ForEach}[1]{\algorithmicforeach\ #1\ \algorithmicdo}

\begin{document}

\title{Deterministic Execution of ROS~2 Applications via Lingua Franca}

\author{%
  \begin{tabular}{@{}c@{\hspace{2.2em}}c@{\hspace{2.2em}}c@{}}
    {\normalsize Harun Teper}
    & {\normalsize Shaokai Lin}
    & {\normalsize Shulu Li} \\
    TU Dortmund University & University of California, Berkeley & University of California, Berkeley \\
    Dortmund, Germany & Berkeley, CA, USA & Berkeley, CA, USA \\
    harun.teper@tu-dortmund.de & shaokai@berkeley.edu & shulu@berkeley.edu \\[1.5em]
    \multicolumn{3}{c}{%
      \begin{tabular}{@{}c@{\hspace{2.2em}}c@{}}
        {\normalsize Edward A. Lee} & {\normalsize Jian-Jia Chen} \\
        University of California, Berkeley & RWTH Aachen University \\
        Berkeley, CA, USA & Aachen, Germany \\
        eal@berkeley.edu & jian-jia.chen@rwth-aachen.de
      \end{tabular}%
    }
  \end{tabular}%
}

% make the title area
\maketitle

\begin{abstract}
The Robot Operating System~2 (\ros{}) is a widely used middleware for robotic systems, characterized by a publish-subscribe (pub-sub) communication mechanism in which computation is structured as callbacks dispatched by \ros{} executors.
Despite its popularity, the pub-sub pattern in \ros{} is inherently nondeterministic: the order in which these callbacks run is nondeterministic even within a single executor, and distributed deployments add further nondeterminism from the interleaving of messages across nodes and from network latency.
Such nondeterminism often leads to concurrency issues and makes it virtually impossible to analyze for safeness and provide guarantees.

We present a framework that is able to convert an unmodified \ros{} application and run it under Lingua Franca (LF), a coordination language for deterministic execution using logical time, so that the same input always produces the same deterministic execution order.
We first describe which \ros{} features can be executed deterministically under logical time.
Such features enable the possibility to establish an automatic conversion framework to extract information from a \ros{} application and directly convert it into an LF program.
The rich features of LF, such as logical-time delays, federated execution across processes, and fault handling, can then be applied to make the \ros{} application be executed in a deterministic and timing-predictable manner without changing the \ros{} code.
We evaluate the framework on a synthetic example and on the Autoware reference system.
We show that the order in which callbacks are executed differs in default \ros{}, while also having end-to-end latencies that vary across executions.
In contrast, our LF-controlled \ros{} system produces a deterministic execution order and consistent end-to-end latencies.

\end{abstract}

\IEEEpeerreviewmaketitle

\section{Introduction}
\label{sec:introduction}
%% Introduction
\ros{}~\cite{ros2} is a widely used middleware for building robotics systems.
It uses the publish-subscribe architecture built on the Data Distribution Service (DDS)~\cite{omg-dds} to enable modular design of distributed systems.
\ros{} has been adopted across different sectors, including autonomous driving, logistics, and industrial automation~\cite{macenski2022robot}.
Some of these applications are safety-critical, which means they require not only timely but also deterministic behavior.

The order in which concurrent events happen on a real machine, however, depends on the operating system, the network, and the hardware.
Without explicit synchronization, that order is nondeterministic, and \ros{} provides no such synchronization in its execution model.
Considerable effort has gone into making the timing of \ros{} systems more predictable~\cite{casini2026survey}.
On the analysis side, Casini et al.~\cite{Casini2019} gave the first response time analysis for \ros{} systems, and subsequent work extended this with response time and end-to-end latency analyses~\cite{Blass2021,blass2021automatic,teper2023timing,Teper2022,teper2024end,tang2023real,tang2024timing}.
On the runtime side, custom executors aim to reduce nondeterminism in practice: PiCAS~\cite{choi2021picas} prioritizes callbacks along causal chains, RTeX~\cite{liu2024rtex} adds lock-free multithreaded scheduling, and the events executor adopted by Teper et al.~\cite{teper2025reconciling} enables nonpreemptive fixed- and dynamic-priority scheduling on top of a FIFO event queue.
These approaches improve predictability, but they do not give the system deterministic execution by construction.

Determinism by construction has been studied in a rich literature on synchronous and dataflow languages~\cite{lee1987synchronous,lee1995dataflow,halbwachs1991lustre,benveniste2003synchronous}, and Lingua Franca (LF) is one such coordination language that has emerged for distributed, concurrent systems.
LF separates \emph{logical time}, the time in which a program reasons about events, from \emph{physical time}, the wall-clock time of the underlying machine, so that any execution of an LF program produces the same logical-time trace.
LF is already used in high-performance robotics middleware research~\cite{kwok2025hprm} and in early industry exploration for safety-critical automotive orchestration~\cite{huang2025sdv}.
\ros{}, however, has a large body of robotics code, libraries, and tooling that LF does not, and porting existing \ros{} systems to LF manually is impractical.

In this paper, we would like to bridge \ros{} applications with the well-studied LF execution paradigm so that we can ensure deterministic execution and achieve timing predictability.
This leads to the central question:

\begin{quote}
\emph{Is it possible to convert \ros{} applications automatically to LF to enable deterministic execution behavior and strong timing guarantees?}
\end{quote}

Our vision is to place an unmodified \ros{} system under the control of the LF runtime, replacing the execution model of \ros{} and its executor with the runtime coordination of LF, enabling deterministic execution by design, while preserving the original code of the \ros{} system.

To the best of our knowledge, the central question above has not been addressed in the literature.
The Lingua Franca Project provides minimal use cases for \ros{} that wrap \ros{}-native code and run it under LF~\cite{lf-playground}, but these are small didactic systems with a single publisher and subscriber, far from an automatic conversion of arbitrary \ros{} applications.

Furthermore, converting any arbitrary \ros{} application into LF is impossible, since some system design aspects of \ros{} cannot be replicated in LF.
For example, \ros{} allows multiple components to send messages to a single message channel, making it impossible for the receiver of said channel to distinguish where the message came from.
This design naturally results in nondeterministic execution and timing behavior, already posing troublesome issues for timing analysis, and is also not possible to be mapped to LF.

As a result, to enable automatic conversion, we have to explore the \ros{} features that admit deterministic execution and ensure the determinism of a \ros{} system through Lingua Franca.
The result of our solution is a conversion framework that extracts the topology of the original \ros{} system and generates a runtime instantiation of the system under LF, without modifying any of the original application code.
\emph{We plan to release the framework as open source for the community to adopt and build upon.}

\textbf{Contributions:} We make the following contributions:
\begin{itemize}
  \item \textbf{Deterministic Subset of \ros{}.}
  In \secref{sec:subset}, we characterize which \ros{} features can be executed deterministically under logical time and which types of \ros{} applications are nondeterministic even under logical time.
  Combining both, we determine the deterministic subset of \ros{} features that we can run deterministically.
  Such characterizations are fundamental to the research question, since it is impossible to enable deterministic and timing-predictable execution if the original semantics of the \ros{} application are naturally unpredictable.

  \item \textbf{Automatic Conversion and LF-controlled Runtime for \ros{}.}
  The features of the deterministic subset of \ros{} enable the possibility to extract information from a \ros{} application and directly convert it into an LF program.
  Towards this, in \secref{sec:subset-tool}, we establish the first automatic conversion framework which extracts the runtime topology of a \ros{} application from its source code and emits an LF program that can run the unmodified \ros{} application code in a deterministic and timing-predictable manner.
  We specify how the LF runtime controls the execution of the \ros{} application and how communication is handled.
  To the best of our knowledge, this is the first framework in this direction.

  \item \textbf{Semantic Enrichment.}
  After the automatic conversion, we can adopt the rich features of LF, such as logical-time delays, federated execution across processes, fault handling, etc., and add them to the converted LF program without modifying the original \ros{} code.
  In \secref{sec:semantics}, we focus on the logical-time delays that allow us to run the system under different latency specifications, and we specify how we use federated execution to enable the concurrent execution of distributed systems in a deterministic and timing-predictable manner.

  \item \textbf{Evaluation.}
  We demonstrate the effectiveness of the framework in enabling deterministic execution and timing predictability for \ros{} applications.
  In \secref{sec:evaluation}, we apply the framework to a synthetic example system and to the Autoware reference system, showing deterministic execution and static ordering of callbacks, a reduction in end-to-end latency variability compared to stock \ros{}, and consistent timer execution behavior.

\end{itemize}

\textbf{Paper Organization.} In addition to the above technical contributions,
\secref{sec:ros2} introduces \ros{}, and \secref{sec:lf} introduces Lingua Franca and its logical-time determinism.
\secref{sec:ros+lf} contrasts the two execution models on a running example, leading to the problem statement in \secref{sec:problem}.

\section{ROS~2}
\label{sec:ros2}

In this section, we introduce \ros{}~\cite{ros2,casini2026survey}, including its basic components and executor model.

A \ros{} system consists of a set of \emph{nodes}, each of which represents one system component.
Each node implements its functionality through a set of \emph{callbacks}, which are methods of the node class.
\ros{} then provides a set of APIs to define how callbacks are triggered, specifying the task types of the system.
There are two types of tasks, time-triggered and event-triggered.
Timers are time-triggered tasks and periodically activated.
Tasks like subscriptions are event-triggered and activated by the arrival of messages on the system's communication channels.

For the communication, \ros{} nodes use \emph{Data Distribution Services} (DDS), a publish-subscribe middleware.
\ros{} defines three forms of communication through DDS: \emph{topics}, \emph{services}, and \emph{actions}.

A \emph{topic} is a publish-subscribe channel.
Any node can define a publisher on a topic, and any callback of that node can use that publisher to send messages.
Subscriptions are defined given a topic, a callback, and a fixed size for its FIFO queue.
When a message is published on a topic, it is broadcast to all subscriptions of that topic, triggering their callbacks.
When a subscriber's callback is executed, it processes the oldest message in its FIFO queue.

In order to run a \ros{} system, the user creates one or more \emph{executors}, which are responsible for scheduling and executing callbacks.
Each node and its callbacks are assigned to exactly one executor, which then manages the execution order of the callbacks that are assigned to it.

\begin{figure*}[!t]
\centering
\begin{minipage}[t]{0.4\textwidth}
\centering
\resizebox{\linewidth}{!}{\input{figures/running_example_ros2}}
\caption{Running example as a \ros{} system.}
\label{fig:running-example-ros2}
\end{minipage}\hfill
\begin{minipage}[t]{0.48\textwidth}
\centering
\includegraphics[width=\linewidth]{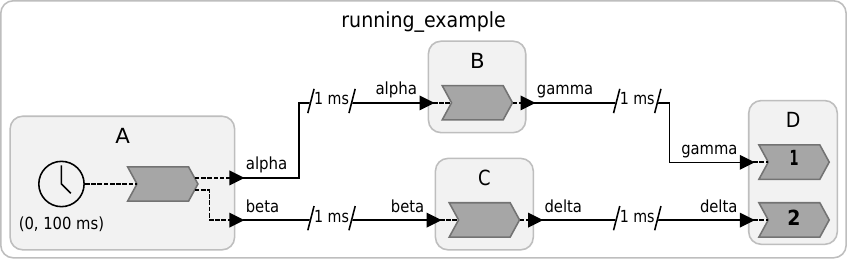}
\caption{The running example expressed in LF.}
\label{fig:running-example-lf}
\end{minipage}
\end{figure*}

Figure~\ref{fig:running-example-ros2} shows the running example used throughout the paper.
It contains four nodes, $A$, $B$, $C$, and $D$, arranged in a diamond and communicating through four topics, \texttt{/alpha}, \texttt{/beta}, \texttt{/gamma}, and \texttt{/delta}.
Node $A$ has a timer with period $T$ that triggers a callback $c_A$ that publishes messages on \texttt{/alpha} and \texttt{/beta}.
Node $B$ has a callback $c_B$ that subscribes to \texttt{/alpha} and publishes on \texttt{/gamma}.
Node $C$ has a callback $c_C$ that subscribes to \texttt{/beta} and publishes on \texttt{/delta}.
Node $D$ has two callbacks, $c_{D1}$ subscribed to \texttt{/gamma} and $c_{D2}$ subscribed to \texttt{/delta}, joining the two paths $A \to B \to D$ and $A \to C \to D$.

\section{Lingua Franca}
\label{sec:lf}
Lingua Franca (LF)~\cite{lohstroh2021determinism,menard2023high} is a coordination language for deterministic concurrent systems.
An LF program describes the structure of a system: which components exist, what triggers them, and how they connect.
Underpinning LF is the reactor model~\cite{lohstroh2020reactors}:
events, such as the execution of system functions in the program, are ordered by \emph{logical time} rather than wall-clock time, which is what lets LF impose a deterministic order on concurrent execution.
\secref{sec:lf-components} introduces the structural building blocks of an LF program, and \secref{sec:lf-time} introduces logical time and the resulting determinism guarantee.
We explain LF by example, using the corresponding system in Figure~\ref{fig:running-example-lf} (as an LF copycat of Figure~\ref{fig:running-example-ros2}) as reference.

\subsection{Reactors, Reactions, and Ports}\label{sec:lf-components}

An LF program specifies the interactions between concurrent components called \emph{reactors}.
Each reactor is a state machine with a fixed set of input ports, output ports, state variables, and reactions.
Reactions are the functions that specify the computation and functionality in an LF system.
Inside of a reactor, reactions can read from and write to the state variables, which encapsulate the state of the reactor.
Reactors can communicate with other reactors only via emitting events on output ports, which then trigger the reactions of the connected input ports of other reactors.
A complete LF program is built by composition.
A top-level reactor instantiates child reactors and wires their input and output ports together, forming the program's dataflow graph.

In our example, illustrated before in Figure~\ref{fig:running-example-ros2}, there are four reactors $A$, $B$, $C$, and $D$, with a connection from $A$ to $B$ via the port \texttt{alpha}, a connection from $A$ to $C$ via the port \texttt{beta}, a connection from $B$ to $D$ via the port \texttt{gamma}, and a connection from $C$ to $D$ via the port \texttt{delta}.

An \emph{event} can be emitted to trigger reactions.
There are four kinds of event sources in LF.
First, messages carried by input/output ports.
Second, LF \emph{timers}, which are periodic event sources declared in a reactor that are triggered at fixed periods.
Third, the built-in triggers \texttt{startup} and \texttt{shutdown} of a reactor, which are triggered once at the first and last moment of the program running.
A fourth source, \emph{actions}, is introduced together with logical time in \secref{sec:lf-time}, together with the time units for each connection in the dataflow graph.

Reactor $A$ has one reaction that is triggered by a timer and writes to the output ports \texttt{alpha} and \texttt{beta}.
Reactor $B$ and $C$ have one reaction each that is triggered by the input ports \texttt{alpha} and \texttt{beta} respectively, and write to the output port \texttt{gamma} and \texttt{delta}, respectively.
Reactor $D$ has two reactions, one triggered by the input port \texttt{gamma} and the other triggered by the input port \texttt{delta}.

\subsection{Logical Time and Determinism}\label{sec:lf-time}

Logical time is the basis for all causal reasoning in an LF program, enabling the deterministic execution of LF programs.
Instead of relying on the physical clock of the system, a central LF runtime maintains a logical clock and advances it according to the events in the program.
The logical time of LF is defined using tags, which are pairs $\tagLF{t}{m}$, where $t$ is a logical timestamp and $m$ is a microstep index.
For an LF system, tags are totally ordered, i.e., for any two tags $g = \tagLF{t}{m}$ and $g' = \tagLF{t'}{m'}$, $g \prec g'$ iff $t < t'$, or $t = t'$ and $m < m'$.
Every event in an LF program carries a tag, whether it is an event generated by an output port, by a timer, or an action.
A reaction triggered at tag $g$ reads only the values present at $g$ and writes outputs at $g$ or later.

An \emph{action} is a reactor element which can be used to trigger reactions at a tag determined by the runtime.
A \emph{logical action} is used to add a logical delay to the program.
The runtime computes the trigger's tag as the current tag plus a programmer-specified delay $d$, and the trigger occurs deterministically at that tag.
A delay of zero at logical time $t$ with a microstep $m$ does not produce the same tag but a tag at the next microstep $m+1$. Hence, chains of zero-delay events are ordered without logical time advancing.
In our example, each connection in the dataflow graph carries a delay through \emph{after delays}, an equivalent form of logical actions that delay events sent over connections.
Each of the four connections carries a logical delay of $1$\,ms, so that the two paths $A \to B \to D$ and $A \to C \to D$ accumulate the same total delay and their events arrive at $D$ at the same logical tag.

The logical time is independent of the physical execution times of reactions by default, forming a well-defined, deterministic timeline of events.
To guarantee determinism for the execution order of multiple reactions inside a reactor, Lingua Franca requires a predetermined static order of all reactions inside of a reactor.
Thus, reactions that are part of the same reactor are run sequentially (\textit{e.g.,} in Figure~\ref{fig:running-example-lf}, the two reactions of reactor $D$ execute in their static order when both are triggered), while reactions that are part of different reactors are independent of each other when run at the same logical time, and can be run concurrently.
Thus, LF derives a determinism guarantee that holds for every program, providing a static order of events if the program is run with the same inputs.

For time-sensitive systems, logical time can be used as timing specifications for the system's physical-time behavior.
In this setting, the logical-time delays introduced by logical actions can represent the execution times, network delays, and deadlines of the system.
The LF runtime then only processes events for which the current physical time is greater than the time value of the logical time tag, i.e., logical time tries to catch up to the physical time.

There is also the possibility to use physical connections in LF, which enables a downstream component to re-timestamp incoming events using the physical time of the system.
This makes the logical timestamps depend on the physical latencies of the system, resulting in nondeterministic execution, as the order of events would depend on the timing of message arrivals, function execution times, and timer firings.

\noindent\textbf{Concurrency and federated execution.}
The LF runtime is responsible for the execution of reactions, using worker threads to run reactions concurrently whenever possible.
In LF, all reactions inside a reactor are guaranteed to be executed sequentially, while reactions of different reactors that are triggered at the same logical time can be executed concurrently, as they are independent of each other.

When extending LF to multi-machine distributed execution, the LF runtime can be implemented as a \emph{federated} runtime~\cite{bateni2023risk}.
Under federation, each machine runs a separate \emph{federate} of the program, which controls a subset of the system's reactors.
The federates are coordinated by the LF Runtime Infrastructure (RTI)~\cite{bateni2023risk}, which maintains the global logical time and ensures the determinism guarantee across machines.
Each federate advances its local logical time independently, but before processing events at tag $g$ it must receive a \emph{Tag Advance Grant} from the RTI, which guarantees that no upstream federate will produce events at tags $\leq g$ in the future.
The determinism guarantee therefore extends from the single-machine setting to the federated setting.

\section{\ros{} vs. LF}
\label{sec:ros+lf}

We now use the running example from Figures~\ref{fig:running-example-ros2} and~\ref{fig:running-example-lf} to illustrate the similarities between the two systems and which components of \ros{} and LF correspond to each other, as well as the differences in their execution models and the resulting nondeterminism in \ros{} that LF closes by design.

Each \ros{} node corresponds to one reactor in LF, the callbacks correspond to the reactions, and each topic connection corresponds to a port connection in LF.
Furthermore, both \ros{} and LF provide a timer to periodically activate the callbacks and reactions, respectively.
The main difference is that the \ros{} system has no logical time components, which are present in the LF system as after delays at each port connection.
In the following, we explain how the actual execution of the systems may differ.

We start with the \ros{} system.
We assume a deployment where each node runs independently in its own process and they communicate over a network via DDS.
Node $A$'s timer activates after its period elapses, executing the callback $c_A$ which publishes on topics \texttt{/alpha} (to B) and \texttt{/beta} (to C).
B's callback $c_B$ receives the message on topic \texttt{/alpha} and publishes on topic \texttt{/gamma} to D, where $c_{D1}$ receives it.
C's callback $c_C$ receives the message on topic \texttt{/beta} and publishes on topic \texttt{/delta} to D, where $c_{D2}$ receives it.

After the callback of node $A$ publishes the messages, the order in which the callbacks $c_B$, $c_C$, $c_{D1}$, and $c_{D2}$ are executed is not deterministic, and depends on the execution times and network latencies of each component.
For example, if the path through $B$ finishes before the path through $C$, then \texttt{/gamma} arrives at D before \texttt{/delta}, even if the developers intend the message of \texttt{/delta} to be processed before the message of \texttt{/gamma} or vice versa.

Next, we move to the LF system, which uses logical-time delays and static ordering inside of the reactor $D$ to provide deterministic execution at every step.
First, after the timer generates an event at the logical time tag $\tagLF{t}{m}$, the reaction of reactor $A$ is executed.
This produces two output events, one on port \texttt{alpha} and one on port \texttt{beta}, both at the same tag $\tagLF{t}{m}$.
Next, due to the after delay of $1$\,ms on the connections from $A$, reactors $B$ and $C$ receive their inputs at the tag $\tagLF{t+1\,\mathrm{ms}}{m}$, and each reactor produces an output on \texttt{gamma} and \texttt{delta}, respectively, at the same tag.
Finally, the two reactions of reactor $D$ are triggered by the outputs of the after delays on \texttt{gamma} and \texttt{delta}, being activated at the same tag $\tagLF{t+2\,\mathrm{ms}}{m}$.
Due to the static ordering of reactions in reactor $D$, the reaction triggered by \texttt{gamma} is executed before the reaction triggered by \texttt{delta}, regardless of the physical execution times and network latencies.
After $100$\,ms, the timer of reactor $A$ generates another event at tag $\tagLF{t+100\,\mathrm{ms}}{m}$, and the same sequence of reactions is executed again, with the same order of execution for the reactions of reactor $D$.

LF produces a deterministic execution order of reactions, and the differences between \ros{} and LF in our running example are as follows:
\begin{itemize}
  \item \textbf{Logical time.} LF uses logical time to order events, while \ros{} relies on physical time, which can lead to nondeterministic execution order due to varying execution times and network latencies.
  \item \textbf{Static ordering.} LF requires a static order of reactions within the same reactor, which ensures a deterministic execution order for reactions triggered at the same logical time.
  While \ros{} provides a fixed-priority order of callbacks for each node, their execution order is heavily dependent on the timing of message arrivals and timer firings, leading to nondeterministic execution ordering.
  \item \textbf{Determinism guarantee.} LF provides a determinism guarantee that holds for every program, ensuring that the same input produces the same execution trace, while \ros{} does not provide such a guarantee, resulting in different execution traces even with the same inputs.
\end{itemize}

\section{Problem Definition}
\label{sec:problem}

After demonstrating the differences between \ros{} and LF and the causes of nondeterminism in \ros{}, we can now turn towards the problem of creating a deterministic version of a \ros{} system by using LF.
For this, we first restate the previous work on the deterministic execution of \ros{} systems and the use of LF for \ros{} systems, and then we define the problem we want to solve in this paper.

Existing work by Bateni et al.~\cite{bateni2023risk} demonstrates the shortcomings of \ros{} that are caused by its nondeterministic behavior.
Specifically, they investigate the autonomous driving software stack Autoware.Auto~\cite{kato2018autoware}, which is built on \ros{}, and observe dangerous out-of-order state sequences that can lead to undefined system behavior and bad data alignment.
First, for multiple data streams that are published from the same component, the order in which other components receive the messages is not fixed.
In Autoware.Auto, one component could receive a velocity signal that does not match the direction in which the car plans to drive, resulting in undefined behavior.
Second, for multi-node systems, the alignment of data from multiple sources is not guaranteed, i.e., if there is a "stop and reverse" signal, and the reverse operation is received and processed without coming to a stop first, then the car may end up trying to reverse at high speed, which can lead to dangerous situations.
To solve this issue, they extracted the application code of Autoware.Auto and re-implemented the components in Lingua Franca manually, demonstrating that the aforementioned issues can be resolved through deterministic execution.
However, this manual conversion is not scalable and does not preserve the original code, making it incompatible with the existing \ros{} ecosystem.

Another line of work by Sagmeister et al.~\cite{rslcpp2025} presents a custom \ros{} executor, called RSLCPP, that enables deterministic execution for single-executor systems.
RSLCPP aggregates all nodes into a single process and drives them through a custom event loop under simulation time, a logical time control mechanism in \ros{}, enabling deterministic execution.
It first determines which timer functions are ready to execute, and then executes all timer functions and all downstream functions that are triggered by the messages sent by the timer functions.
This is repeated until there are no more functions to execute, at which point the logical time is advanced to the next point in time when a timer function is ready to execute.
This approach achieves determinism in \ros{}, but it comes with significant structural restrictions.
First, it is restricted to a single process, since the coordination between multiple processes under simulation time poses significant challenges regarding the synchronization of logical time across processes and the ordering of events across processes.
Second, the authors cut off communication to all components that are not run by the RSLCPP executor, since those could introduce nondeterminism through their interactions with the components run by RSLCPP.

Interestingly, the authors of RSLCPP also identify that certain design patterns of \ros{} systems cannot be included in their approach, such as blocking functions, which can cause the entire system to stall and thus break the logical time control of the RSLCPP executor.
Furthermore, they also explore and implement the idea of delay modeling for execution times of callbacks, which introduces logical time offsets for the callback execution.
This allows them to analyze the system behavior under varying latency conditions, while staying independent of the actual runtimes of the functions.

These works demonstrate the importance of deterministic execution for \ros{} systems, and the potential of LF to achieve it.
However, they are limited regarding their usability, expressiveness, and compatibility with the existing \ros{} ecosystem.
In order to bring deterministic execution to \ros{}, we aim to answer the following questions:
\begin{itemize}
  \item \textbf{Q1.} \emph{Which \ros{} features admit deterministic execution, and which cannot?}
  \secref{sec:subset} defines the deterministic subset of \ros{} and the features that fall outside it.
  \item \textbf{Q2.} \emph{Given a \ros{} system built entirely from the deterministic subset of \ros{}, can we construct the corresponding LF program automatically from the source?}
  \secref{sec:subset-tool} and the Appendix answer this with an extraction-and-generation pipeline that produces an LF program running the unmodified \ros{} code.
  \item \textbf{Q3.} \emph{What execution-time semantics can be added and adjusted through LF to the converted system on top of deterministic callback ordering?}
  \secref{sec:semantics} details the LF features the converted system may use and their effect on its behavior.
\end{itemize}

\section{Deterministic Subset of \ros{}}
\label{sec:subset}

In this section, we characterize the subset of \ros{} that admits deterministic execution.
For this, we first discuss what currently prevents \ros{} from coordinating the execution of its components in a deterministic way, and then we explain what is needed to enable determinism, and which parts of \ros{} cannot be included in the deterministic subset.

\subsection{Nondeterminism in \ros{}}
\label{sec:subset-nondeterminism}

\ros{} systems are nondeterministic by design in (at least) three identifiable ways.

\noindent\textbf{(A) Callback execution ordering.}
The execution of callbacks in \ros{} solely depends on the elapsing of timer periods and the arrival timing of messages.
When several callbacks arrive at roughly the same time, the \ros{} executor
simply starts executing as soon as the first callback arrives and is ready, resulting in nondeterminism.

\noindent\textbf{(B) Multi-component coordination.}
For a system with multiple components, the sequence in which those components are executed and how data propagates through the system depends on the execution times and network latencies, causing nondeterminism in the order of execution across components and the alignment of data from different components.
The discussion in \secref{sec:ros+lf} already demonstrates this, where we have two different data paths that arrive at a final component, whose execution order depends on when the data arrives, resulting in nondeterministic behavior.

\noindent\textbf{(C) Unsynchronized timers.}
\ros{} timers should be executed periodically. However, two problems arise due to the design of \ros{} timers.
(i) Their behavior depends on the execution times of other callbacks, which can delay the timer execution due to the nonpreemptive scheduling behavior of \ros{}.
(ii) For multiple components, there is no mechanism to synchronize timers or control the offset at which they execute relative to one another.
For example, if there are two timers with the same period, \ros{} provides no mechanism to control their offsets.
Instead, the offset simply depends on the timing of when the timers are created, which depends on startup times and system delays, causing misaligned timer executions.

The common cause underlying (A), (B), and (C) is that \ros{} ties scheduling decisions to physical time.
Callback ordering depends on which message a thread picks up first, multi-component coordination depends on execution times and network latencies, and timer offsets depend on startup delays.
A scheduler that advances time independently of how fast the hardware runs and that fixes the order of simultaneous events at the program level would remove all three sources at once.

\subsection{Determinism via Logical Time}
\label{sec:subset-determinism}

Determinism requires two concepts that \ros{} lacks by default.
The first is a discrete time model independent of the physical clock, so that the schedule is a property of the program rather than of the hardware running it.
The second is a deterministic ordering rule for events that share an instant in that time model.
As shown by Sagmeister et al.~\cite{rslcpp2025}, the simulation time of \ros{} provides a discrete clock that components can follow.
However, their solution did not yet resolve coordinating multiple components and arbitrating what should run when multiple events occur simultaneously.

Lingua Franca supplies both concepts needed for deterministic execution directly through logical time, as described in \secref{sec:lf-time}.
Using tags $\tagLF{t}{m}$, LF ensures a total order of events across the entire program, and advances time only when every reaction at the current tag has produced its outputs.
This decouples the schedule from physical execution times so that an LF program always has the same sequence of tags regardless of how fast the hardware happens to be.
At each tag, reactions inside a reactor execute in a static order declared in the reactor, while reactions in different reactors at the same tag are unordered and may run concurrently.
However, once events with the same tag from different components arrive at the same reactor, the static order of reactions inside that reactor determines the execution order of those events.
This enables developers to control both the order of events inside a component, as well as the order in which data from different components is processed.

Logical time also serves as a means of modeling time-sensitive systems.
Logical-time delays on connections (\emph{after delays} in LF) can stand for execution times, network latencies, and deadlines, making the runtime only advance time once the corresponding physical time has elapsed.
This way, the schedule remains a property of the program rather than of the hardware, even when the program describes real-time behavior.

Under this model, the three sources from \secref{sec:subset-nondeterminism} dissolve.
(A) Reaction order inside a reactor is statically declared, i.e., two callbacks on the same node execute in the order the source defines.
(B) Tag-ordered delivery replaces latency-dependent arrival of events.
Thus, all components wait for the logical clock rather than for the wall-clock to schedule callbacks, making coordination deterministic across components.
(C) Two logical timers with the same nominal period fire at the same sequence of tags throughout the program.
Their relative offset can be configured and is fixed, making them deterministic by design.

We can define the deterministic subset of \ros{} as the set of \ros{} systems that only use the features which admit deterministic execution under a logical-time scheduler.

\subsection{Nondeterminism Even under Logical Time}
\label{sec:subset-blocking}

Some \ros{} features remain nondeterministic even under a logical-time scheduler, since \ros{} provides a lot of freedom regarding the system design.
In the following, we briefly present some of the features that can be used in \ros{} but cannot be included in the deterministic subset.
To the best of our knowledge, those features have not been included in any prior studies on real-time systems and \ros{}, since they introduce behaviors that are difficult to analyze.

\begin{itemize}
\item
\noindent\textbf{Multiple publishers on one topic.}
When multiple DDS publishers share a topic, the order in which their messages arrive at a subscriber depends on the execution order and network delay of the message senders.
For example, take two publishers $p_1, p_2$ on topic \texttt{/state} and one subscriber $s$ on the same topic, where each publisher publishes at period $T = 100$\,ms.
At the subscriber, the delivery sequence may be $(p_1, p_2, p_1, p_2, \ldots)$ on one run and $(p_2, p_1, p_2, p_1, \ldots)$ on another, depending on OS scheduling and network load.
As \ros{} uses a FIFO queue for subscriptions, there is no way to control in which order the messages in the buffer are processed.
Thus, the order of messages from multiple publishers on the same topic is nondeterministic, and we restrict ourselves to systems where each topic has at most one publisher.
This is also a common assumption in existing \ros{} analyses, which either assume a DAG structure or at most one publisher per subscription~\cite{casini2026survey}.

\item
\noindent\textbf{Blocking service clients.}
\ros{} callbacks are intended to be non-blocking functions.
However, if developers use blocking functions inside their callbacks, for example by calling a service client synchronously, the calling executor thread stalls until the response arrives, preventing the advancement of logical time, and stalling the complete system.
This was also excluded by Sagmeister et al.~\cite{rslcpp2025} for deterministic execution.

\item
\noindent\textbf{Dynamic system structure.}
The structure of a \ros{} system can change at runtime in ways that are not predetermined.
This contradicts the static structure of the underlying reactor model of LF.
Existing work on \ros{} systems assumes that the system and the resulting model are fixed and do not change at runtime~\cite{casini2026survey}.

\end{itemize}

\section{LF-based Execution of \ros{} Systems}
\label{sec:subset-tool}
After determining the set of features that admit deterministic execution, we now turn to the problem of running a \ros{} system using Lingua Franca, and how we can enable LF-based execution of \ros{} systems without modifying the original application code.
For this, we first present how we convert a \ros{} system into an LF program automatically, and how the generated program preserves the behavior of the original system while providing deterministic execution.
Then, we present how we integrate the LF runtime to control the execution of \ros{}-native code without any changes, and how we handle communication for data transport and scheduling.

\subsection{Automatic Conversion of \ros{} to LF}\label{sec:subset-tool-conversion}
The first part is the automatic conversion of a \ros{} system into an LF program.
For this, we have built a tool that reads the source code the user provides, and extracts a static system graph that captures every node, callback, timer, publisher, subscription, and topic connection.
It extracts the information from the launch files of the \ros{} system, which specify which nodes are part of the system and what parameters are used.
Furthermore, we use static code analysis for the C++ source code of each node, which provides the details of the callbacks and what functions they call, such as publish and subscribe functions, and what variables they write to and read from.
Finally, we also extract the CMake metadata of a built \ros{} workspace, which provides information about how the nodes are built and connected together.
The extracted graph captures the structure of the system, the callbacks and their interactions, and how the system is built and connected together, which are all necessary for the conversion to LF.

From the graph, the tool emits \texttt{.lf} source files that mirror the structure of the original system.
For example, every node is a reactor, every callback a reaction, and all topic connections are port connections in the generated LF program.
The LF runtime can then use the LF program to determine the order for the execution of the callbacks.
Importantly, the generated program preserves the callbacks of the original \ros{} program, their connections, but adds the possibility to control their execution order using logical time.

Please note that this tool may not successfully convert every system, as described in \secref{sec:subset-blocking}, but it can convert any system that is built using the deterministic subset of \ros{}.
The Appendix describes the framework in detail.

\subsection{Lingua Franca Runtime for \ros{}}\label{sec:subset-tool-runtime}
Next, we present how to integrate the LF runtime to run \ros{}-native code using the LF runtime instead of the \ros{} executor.
For this, we consider two main designs, either (1) to run the original callbacks in a \ros{} executor that coordinates with the LF runtime, or (2) to run the callbacks directly using the LF runtime.

For option (1), this would entail that there are both LF processes and \ros{} processes running together, and the LF runtime would have to coordinate with the \ros{} executors when to execute which callbacks.
This coordination between the processes poses great challenges for the deterministic execution of the system.
First, it would still rely on \ros{} executors running the callbacks, which would require changes to their scheduling behavior to control the execution order and make it deterministic.
Second, the LF runtime would need to receive updates from the executors to know when the callbacks are executed and when they are done.
This would require additional communication and synchronization to guarantee deterministic execution.

Option (2) directly runs the original callbacks using the LF runtime.
This design is simpler, since there is only one runtime that is processing everything, making it possible to directly execute the callbacks without needing to coordinate with a separate \ros{} executor.
Furthermore, concurrency in the program is directly handled by the LF runtime, which automatically decides which callbacks can run simultaneously.
Thus, we opt for this option in our final design.

We note that for Option (2), the LF runtime creates one executor instance to store references of all system nodes and their callbacks.
Importantly, that executor is not running at all and does not control any part of the execution of callbacks.
The LF runtime uses the executor and a binding function to directly call and execute the unmodified \ros{} callbacks.

\subsection{Communication and Coordination}
To run the \ros{} system under LF control, we need to manage the communication between the components of the system.
There are two types of communication we need to consider.
First, the data transport between the components, which is originally handled by DDS in the \ros{} system.
Second, the event transport between the components in the LF program, which generates the events that are used to ensure that the execution order of the callbacks is deterministic and follows the logical time semantics of LF.

For the data transport, since we do not want to modify any application code, we leave the \ros{} callbacks untouched.
Therefore, they still exchange data with other components via DDS.
Furthermore, this design preserves the ability for data to be received by \ros{} components that are not controlled by LF, e.g., debugging, data logging, and visualization tools that are part of the \ros{} ecosystem, but do not necessarily need to be run in a deterministic way.

For the event transport, we mirror the structure of the original system in the generated LF program, including the connections between reactions.
These connections then use dummy LF events that are needed only to encode the corresponding logical times to control the deterministic ordering of the system.
The event messages are minimal, so they have negligible overhead.

\section{Adding LF Semantics on Top}
\label{sec:semantics}

Once a \ros{} system is realized under the LF runtime, the runtime can enforce execution-time semantics that \ros{} has no equivalence for, such as logical-time delays, custom event ordering, and fault handling.
In the following, we detail how we augment the converted system with some LF semantics and discuss how they affect the system behavior.

\subsection{After Delays and Logical Actions}
\label{sec:semantics-after}

Each connection in the generated LF program may carry a logical-time delay (also known as an after delay) that represents the time that passes for execution and network operations.
We use the connections between ports to insert those delays, representing the time for the callback to run, and the DDS message to pass from the publisher to the subscription.
Just like in RSLCPP~\cite{rslcpp2025}, the delays allow \textit{defining} the system behavior under varying latency conditions.
Through the logical delays, system timing is a first-class, statically configurable property of the program
rather than just an observed quantity at runtime.

For the choice of the delays, different approaches are viable, depending on whether consistency (i.e., ensuring that the system agrees on the shared information) or availability (i.e., ensuring that the system reacts in time when some data is not available) is more important for the system, as described by the CAL theorem (\textbf{c}onsistency, \textbf{a}vailability, \textbf{l}atency)~\cite{lee2021cal,lee2023cal}.

When ensuring strong consistency, as a first option, one would not set a logical delay to the connection, rendering
the upstream sending event and the downstream receiving event logically instantaneous.
When the downstream reaction is invoked, the reaction blocks until the underlying payload is delivered over DDS.
With the absence of a logical delay, when the upstream component sends a message at timestamp $t$,
the downstream component cannot process its internal events with timestamps greater than $t$.
Thus the downstream component moves in sync with the upstream component
and always agrees on the shared data value at the \textit{same} timestamp $t$ regardless of the physical latencies incurred, achieving strong consistency.

A second option favoring strong availability would use a \textit{physical connection}, which enables the downstream component to re-timestamp an incoming message using its physical clock, making the timing of the system dependent on the physical latencies strictly.
By replacing the sending timestamp with the receiving component's physical clock value, the physical connection removes the monotonicity constraint of the messages sent, effectively falling back to the pub/sub semantics of \ros{}, which delivers maximum availability.

A third option landing a trade-off between the two extremes is to set the logical delay to the maximum observed latency between the upstream and the downstream components.
This produces schedules whose logical time is guaranteed to be no faster than the physical time, resulting in behavior where the upstream and the downstream suffer \textit{bounded} inconsistency, while preserving some degree of availability.

Yet another option is to set the latencies to the mean observed latency, which produces schedules that are faster than the physical time in average, but can lead to some messages being late.
Through LF, one can then detect late messages, and add custom fault handling mechanisms, such as warnings, recovery code, or even stopping the program, without modifying the original \ros{} code.

In total, the choice of the logical delays affects the logical time steps of events and thus also the physical time it takes to process data along the tasks.
Furthermore, the logical-time delays can be used to extend the semantics of the system with custom fault handling for late messages, without modifying the original code.

\subsection{Federated Coordination}
\label{sec:semantics-federation}

As \ros{} systems are distributed, we employ LF's federated execution model to coordinate the execution of multiple components across processes and machines.
Federation allows multiple LF programs, called federates, to run concurrently and coordinate their execution through a shared logical time.
The federates communicate through the LF Runtime Infrastructure (RTI)~\cite{bateni2023risk}, which ensures that the logical time of each federate advances in a way that respects causality across the entire system.
As a result, the use of federation extends the determinism guarantee from a single process to a distributed multi-process deployment.

\section{Evaluation}
\label{sec:evaluation}

We apply our automatic conversion framework to two \ros{} systems to run them using the LF runtime, showing that the converted systems have deterministic execution order of callbacks, consistent end-to-end latencies, and consistent timer execution behaviors.

The first system is the Autoware reference system~\cite{autoware-ref}, a synthetic benchmark that replicates the autonomous driving pipeline of Autoware~\cite{kato2018autoware}, including $24$ nodes, $23$ topics, and $39$ causal connections spanning perception, localization, planning, and control.
The second system is our running example based on Figure~\ref{fig:running-example-ros2}, which we use to demonstrate the effects of different logical-time delays on the execution behavior, and the trade-offs between consistency and availability.

\subsection{Setup}
\label{sec:eval-setup}

Experiments are run on \ros{} Jazzy in a Docker container on an Ubuntu~$22.04$ host with Linux kernel~$6.8$, using an AMD Ryzen~9~5900X ($12$ cores / $24$ threads), and $32$\,GB of RAM.
For the native \ros{} execution, all nodes are launched as separate processes using the default \texttt{SingleThreadedExecutor}.
For the DDS communication, we use the default \texttt{rmw\_fastrtps\_cpp} implementation.
The LF-based execution uses Lingua Franca~$0.12.1$~\cite{lingua-franca}.
We use \texttt{ros2\_tracing} to capture \ros{} tracepoints, including the start times of callbacks to reconstruct per-callback invocations and their causal ordering.
We run each experiment configuration twenty times for $60$\,s per run.
\emph{We intend to release the source of the conversion framework, the two systems under test, and the experiment scripts.}

\subsection{Autoware reference system}
\label{sec:eval-refsys}

The Autoware reference system includes $24$ nodes that run as separate processes communicating over DDS, so nondeterminism from DDS arrival timing, executor races, and timer phase differences is expected to occur.
We focus on the full LiDAR pipeline~\cite{autoware-ref}, in which ten nodes process data originating from a LiDAR sensor to produce control signals for the actuators of the vehicle.

For the end-to-end latency, we measure the time from the moment the LiDAR starts executing until the moment the corresponding control signal is emitted by the \texttt{VehicleInterface} node.
There are two timers with a period of $100$\,ms in the data path, one in the \texttt{FrontLidarDriver} node that produces the LiDAR messages, and one in the \texttt{BehaviorPlanner} node that produces a plan based on the LiDAR data and other inputs.
Per-callback execution times are implemented as a fixed amount of arithmetic work, set to roughly $0.2$\,ms of execution time per callback on our platform.

Table~\ref{tab:refsys} reports the metrics across $20$ runs of $60$\,s each, both over the full run and over the steady state after discarding the first $5$\,s.
The first seconds are affected by a startup transient, as DDS discovery and \ros{} node creation complete only after the program begins.
Our framework removes this transient from the LF execution by anchoring logical time zero to the instant all subscriptions have matched a publisher, which keeps the logical ordering unchanged but aligns the first events' physical-time release with the logical schedule.
Default \ros{} does not have such a mechanism.

\begin{table}[t]
\renewcommand{\arraystretch}{1.2}
\caption{Autoware reference system: per-runtime metrics, $N = 20$ runs of $60$\,s each, over the full run and over the steady state.}
\label{tab:refsys}
\centering
\small
\begin{tabular}{@{}lrr@{}}
\toprule
\emph{Metric} & \emph{Default \ros{}} & \emph{LF runtime} \\
\midrule
Pair order drift across runs    & up to 100\%   & 0\%        \\
Pairs with $>$10\% order drift  & 12 / 15       & 0 / 15     \\
\midrule
\multicolumn{3}{@{}l}{\emph{Full $60$\,s run}} \\
E2E mean [ms] (per run)         & 5.4 -- 93.6   & 100.8 -- 100.9 \\
E2E std [ms] (per run)          & 15.4 -- 130.6 & 0.22 -- 0.41   \\
E2E p99 [ms] (worst run)        & 866.1         & 101.7          \\
E2E max [ms] (worst run)        & 1\,455.1      & 102.1          \\
\midrule
\multicolumn{3}{@{}l}{\emph{Steady state ($5$\,s discarded)}} \\
E2E mean [ms] (per run)         & 2.8 -- 87.4   & 100.8 -- 100.9 \\
E2E std [ms] (per run)          & 0.06 -- 25.8  & 0.22 -- 0.42   \\
E2E p99 [ms] (worst run)        & 101.7         & 101.7          \\
E2E max [ms] (worst run)        & 101.8         & 102.1          \\
\bottomrule
\end{tabular}
\end{table}

\noindent\textbf{Within-node callback ordering.}
We focus on \texttt{BehaviorPlanner}, which subscribes to six topics (\texttt{NDTLocalizer}, \texttt{ObjectCollisionEstimator}, \texttt{Lanelet2GlobalPlanner}, \texttt{Lanelet2MapLoader}, \texttt{ParkingPlanner}, \texttt{LanePlanner}) and outputs a plan when its $100$\,ms timer executes.
For each of the fifteen pairs of subscription callbacks and each $100$\,ms timer interval between the planner executions, we record which subscription fires first.
We then investigate, for each callback pair, whether the order is the same across runs.

Under default \ros{}, the order between two subscriptions changes from run to run.
For twelve of the fifteen pairs, how often one subscription fires before the other shifts by more than ten percentage points between runs.
For three pairs the dominant order flips entirely.
One run has sub $A$ firing before sub $B$ in nearly every window, while another run of the same system has $B$ firing before $A$ in nearly every window.
The same system processing the same workload thus processes data in different sequences, resulting in the planner using different information to generate its output on every run.

Under LF the order of every pair is identical across all twenty runs.
Each subscription callback fires at a deterministic logical-time tag, and ties at the same tag are resolved by the static reaction order inside the reactor.

\noindent\textbf{End-to-end latency.}
In the steady state, both runtimes reach a comparable worst case, a p99 of $101.7$\,ms and a maximum near $102$\,ms.
What differs is the mean and its reproducibility.
Under LF the mean is $100.8$--$100.9$\,ms across all runs with a within-run standard deviation of $0.22$--$0.42$\,ms, whereas under default \ros{} the mean differs on every run, ranging from $2.8$ to $87.4$\,ms.
Over the full run, the LF statistics are unchanged because no chain-job is delayed by the startup transient, while default \ros{} reaches a p99 of $866$\,ms and a maximum of $1\,455$\,ms from the chain-jobs that run during discovery.
Both runtimes thus reach the same steady state, but only LF reaches it without a large startup transient.

The dominant component of the steady-state chain latency is caused at the \texttt{BehaviorPlanner} node.
The \texttt{BehaviorPlanner} processes LiDAR data only when its own $100$\,ms timer fires.
A LiDAR sample that arrives at the planner gate must therefore wait until the next timer execution before the data is processed.
The duration of this wait depends entirely on the relative phase of the two $100$\,ms timers along the chain.
Under LF, both timers fire at the same sequence of tags from logical time $T = 0$ onwards (\secref{sec:subset-determinism}), resulting in a consistent delay of about $100$\,ms on every run for the end-to-end latency.
Under default \ros{}, the two timers run on independent physical clocks, and the phase offset between them is whatever it happened to be when each node finished its initialization.
The delay introduced by waiting on the timer is therefore a run-specific constant somewhere in the range of $[0, 100)$\,ms, producing a different value on every run.

\subsection{Determinism on the running example}
\label{sec:eval-runex}
In the following, we show how we can use the running example (Figure~\ref{fig:running-example-ros2}) to demonstrate the effects of different logical-time delays on the execution behavior.
The running example is instantiated with a timer period of $T = 100$\,ms.
To expose the nondeterministic execution order at Node $D$ identified in \secref{sec:subset-nondeterminism}, the execution times of the two paths $A \to B \to D$ and $A \to C \to D$ are designed to be comparable but varying over runs.
Specifically, Node $A$'s timer callback $c_A$ publishes on \texttt{/alpha} and \texttt{/beta} after a fixed $1$\,ms of execution time.
The parallel paths via $B$ and $C$ propagate the data to Node $D$.
Node $B$'s callback $c_B$ runs for a random execution time drawn uniformly from the interval $[0, 10]$\,ms before publishing on \texttt{/gamma}, and Node $C$'s callback $c_C$ runs for an independent random execution time drawn from the same interval $[0, 10]$\,ms before publishing on \texttt{/delta}.
Node $D$'s callbacks $c_{D1}$ and $c_{D2}$ each run at a fixed $1$\,ms of execution time.

The arrival order of \texttt{/gamma} and \texttt{/delta} at $D$ in \ros{} due to the above design varies over time.
We measure two quantities on Node $D$ across $20 \times 60$\,s runs per configuration.
First, the fraction of timer ticks for which $c_{D1}$ starts before $c_{D2}$.
Second, the end-to-end latency from $A$'s timer start to the later of $c_{D1}$'s and $c_{D2}$'s end for the same tick.

\smallskip
\noindent\textbf{Ordering.}
Under default \ros{}, $c_{D1}$ precedes $c_{D2}$ on $49.0\%$ of executions, showing the randomness of the order at $D$.
The randomly varying work for the reactions in reactors $B$ and $C$ makes the two paths to $D$ finish in a different order from one execution to the next, and the \ros{} executor on $D$ serves whichever message arrived first.
Under LF, every configuration places $c_{D1}$ before $c_{D2}$ on all ticks, since their logical arrival times at $D$ are identical by construction, and the static reaction order inside $D$ resolves ties in favor of reaction $c_{D1}$.

\begin{figure}[t]
\centering
\includegraphics[width=\linewidth]{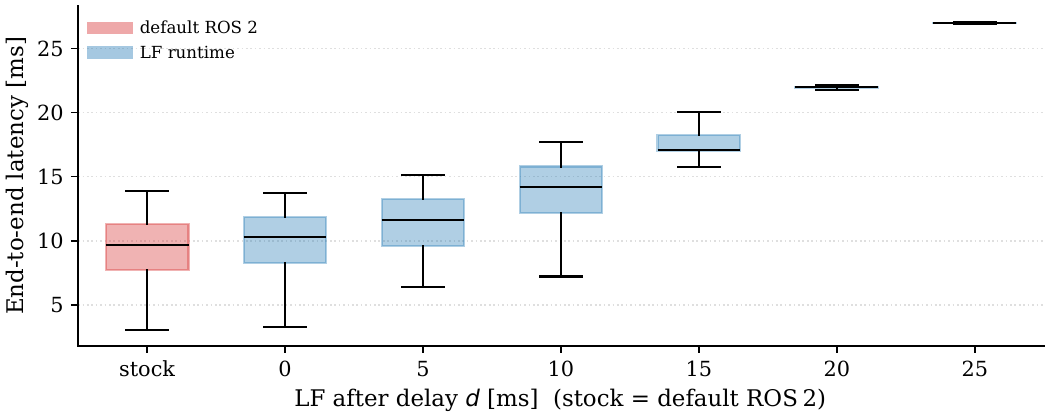}
\caption{Running example: end-to-end latency distribution under default \ros{} and LF across the after-delay sweep ($N = 20$ runs of $60$\,s each). Boxes show the interquartile range, whiskers the $5$th--$95$th percentiles.}
\label{fig:runex-sweep}
\end{figure}

\noindent\textbf{The after delay as a latency-consistency knob.}
We study the effects of logical-time delays on the execution behavior of the system, and how logical delays can be used as a design knob to control the consistency and availability of the system.
For this, we choose different values for the logical-time delays of the connections while preserving the constraint that both paths from $A$ to $D$ carry the same total logical delay $d$: $d/2$ on each of the four edges $A \to B$, $B \to D$, $A \to C$, and $C \to D$.
Figure~\ref{fig:runex-sweep} reports the default \ros{} baseline against LF configurations sweeping $d$ from $0$ to $25$\,ms in steps of $5$\,ms.

The median end-to-end latency rises from $\approx\!10$\,ms at $d = 0$ to $\approx\!27$\,ms at $d = 25$\,ms, gaining roughly $5$\,ms for every $5$\,ms added to $d$, on top of a workload floor of a few milliseconds.
The spread of the distribution tells a more interesting story.
At $d = 0$ the standard deviation of the end-to-end latency is about $2.4$\,ms, because the two paths propagate their data randomly to the output without any controlled delay.
This setting maximizes consistency, since both paths deliver the freshest data the moment it is ready.
As $d$ grows, the logical delay progressively absorbs the temporal variability of the two paths, and the boxes tighten until, at $d = 20$\,ms and beyond, the standard deviation collapses below $0.1$\,ms because both reactions are triggered at the same fixed tag on every tick.
This setting maximizes availability, since the latency becomes a constant the rest of the system can rely on, at the cost of no longer using the most up-to-date data.
The after delay therefore trades consistency for availability, and the designer can configure this aspect through the logical-time delays of the connections.
Default \ros{} does not offer such control, and its latency and ordering are dependent on when the executor finishes callbacks and DDS delivers messages.

\section{Conclusion}
\label{sec:conclusion}

We characterize the subset of \ros{} features that admits deterministic execution under Lingua Franca, and build a tool to convert any \ros{} application in this subset into an LF program automatically.
The generated program runs the unmodified \ros{} callbacks under the LF runtime, which takes over scheduling and callback ordering to guarantee deterministic execution.
On top of this conversion, the system gains LF-native mechanisms for logical-time delays, federated coordination, and fault handling, without any change to the application code, which can be used to control the consistency and availability of the system.
Evaluation on the Autoware reference system and a synthetic scenario shows deterministic callback ordering and controlled end-to-end latency variability.

\appendix
\label{sec:appendix-tool}

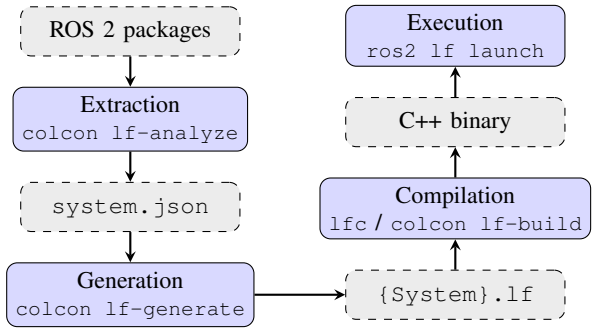
\begin{figure}[t]
\centering
\input{figures/pipeline}
\caption{The \texttt{colcon~lf} pipeline: launch files, C++ source, and CMake metadata are extracted to \texttt{system.json}, then converted to \texttt{.lf} source by applying the embedding $\embed$, compiled by \texttt{lfc}, and executed under LF time authority.}
\label{fig:pipeline}
\end{figure}

In the following, we detail the extraction and conversion tools that we use to automatically convert a \ros{} system into an LF program and run it, as illustrated in Figure~\ref{fig:pipeline}.

\subsection{Extraction}
\label{sec:appendix-tool-extract}

We start from the \ros{} code from which we extract the system information.
We first describe how \ros{} code is organized in a codebase, and how we extract the necessary information from it to build a system graph that captures the structure of the system and the components' interactions.

\ros{} provides so-called packages in which the source code of a \ros{} system is organized.
A package is a directory that contains the source code of said package, configuration files, and any resources necessary to run the system.
Each package specifies its libraries and executables in a \texttt{CMakeLists.txt} file, which is used to build the package and link the source code together.
Furthermore, each package may include launch files, which specify the executables to run and their parameters.
Launch files can also be nested to include launch files from multiple packages.

Our extraction tool is built on top of the \texttt{colcon} build system, which is the standard build system for \ros{}, and we call this step \texttt{colcon~lf-analyze}.
It parses all packages in a workspace, and extracts the necessary information from the launch files, source code, and CMake metadata using static code analysis.
Since we only require an instantiation of the static system for conversion, static code analysis is sufficient for the extraction.
If the extracted system contains any feature that prevents deterministic execution, as described in \secref{sec:subset-blocking}, the tool refuses to generate a program and provides feedback on which features are blocking the conversion and which part of the code includes them.

The output of the extraction tool is a \emph{system graph} written to a single \texttt{json} file, enumerating all components that are extracted, including launch files, executables, libraries, callbacks, timers, publishers, subscriptions, and so on.

\subsection{Generation}
\label{sec:appendix-tool-gen}
Next, we take the system graph and generate \texttt{.lf} source files that mirror the structure of the original system.
This step is called \texttt{colcon~lf-generate}.
The trivially convertible \ros{} components map one-to-one to LF features as follows:

\begin{itemize}
  \item \textbf{Nodes:} Each node maps to an LF reactor.
  \item \textbf{Callbacks:} Each callback of a node maps to a reaction of the reactor.
  \item \textbf{Timers:} Each timer with a fixed period $T$ maps to an LF timer with the same period.
  \item \textbf{Publishers:} Each publisher maps to an output port.
  \item \textbf{Subscriptions:} Each subscription maps to an input port.
  \item \textbf{Topics:} A topic determined by the unique publisher and subscription pair corresponds to the connection between an output port and input port.
\end{itemize}

Next, we move on to the more complex features that have a non-trivial LF realization, but still can be converted.

\noindent\textbf{Topics with multiple subscribers.}
A topic with multiple subscribers maps to one output port connected to multiple input ports.

\noindent\textbf{Async service clients.}
A service client and a service server are usually implemented as two nodes, where a client sends a request to a server, and the server sends a response back.
In the node hosting the service client, one callback is responsible for sending the request, and another callback is responsible for processing the response.
In the node hosting the service server, one callback is responsible for processing the request and sending the response back.
Our tool maps each of the three callbacks to its own reaction, and adds two connections: client-request to server, and server-response to client.

\noindent\textbf{Mutually exclusive callback groups.}
\ros{} provides the concept of callback groups, which are a way to group callbacks together and specify if they can be executed concurrently.
Mutually exclusive callback groups prevent the concurrent execution of all callbacks that are part of the group, while callbacks of different mutually-exclusive callback groups are allowed to run concurrently.
By default, each node has one mutually-exclusive callback group.
If not specified otherwise, each callback is part of the default callback group of its node.

In LF, all reactions inside a reactor are guaranteed to be executed sequentially.
Thus, we map callback groups as nested callback group reactors inside node reactors.
This reproduces the concurrency of \ros{} in LF, as reactions from different callback group reactors can run concurrently, while the reactions in one callback group reactor are run sequentially.

\noindent\textbf{Statically known topic names, namespaces, and remappings.}
All naming schemes, such as for nodes, topics, callbacks, can be extracted from the \ros{} system under the assumption that these do not change during runtime.

While there are many more features in \ros{}, such as lifecycle nodes, actions, and reentrant callback groups, that are used in practice, we focus on the previously described features, as they are sufficient to cover the conversion of a large class of \ros{} systems to LF programs.

\subsection{Compilation}
\label{sec:appendix-tool-compile}

The generated \texttt{.lf} sources are compiled by the LF compiler \texttt{lfc}, which emits C++ reaction code that links against the LF runtime.
We refer to this step as \texttt{colcon~lf-build}, which builds the resulting binaries together with thin C++ wrapper packages and new launch files that connect the LF executables to the original \ros{} code.

\noindent\textbf{Runtime binding.}
The generated program runs under the LF runtime, which schedules reaction firings according to the logical time semantics of the LF runtime.
A thin C++ library, \texttt{ros2\_lf\_binding}, bridges the two runtimes.
Each generated reaction calls a corresponding function, e.g. \texttt{ros2\_lf\_invoke\_timer} for timers or \texttt{ros2\_lf\_invoke\_sub} for subscriptions, which dispatches into the unmodified \ros{} callback body through the registered handle.
The \ros{} runtime threads are not started.
Instead, the binding extends the default \ros{} executor with a function that runs a single callback, which keeps DDS message reception and other built-in \ros{} services available, while the program is under direct control of the LF runtime.

\noindent\textbf{Communication and time.}
Message payloads continue to flow over DDS exactly as in the original program, since the unmodified callback bodies still call \texttt{publish()} and \texttt{take()} for DDS communication.
On the LF side, the replicated system sends only dummy messages between ports, because the LF runtime is used solely to coordinate activation and ordering, not payload transport.
The \ros{} clock is wired to logical time, and a custom \ros{} publisher inside the LF runtime emits the current logical time on a dedicated topic using a custom message type that includes both the logical time and the microstep data.
All original \ros{} timers are kept inactive, since the LF runtime controls the triggering of those timers.

\subsection{Execution}
\label{sec:appendix-tool-exec}

The final step in the pipeline launches the compiled program under the LF runtime.
Invoking \texttt{ros2 lf launch} starts the binaries produced by \texttt{colcon~lf-build} using the generated launch files and registers the binaries as standard \ros{} executables, so external \ros{} tools such as \texttt{ros2 topic}, \texttt{ros2 bag}, and \texttt{rqt} continue to work with the running system without any modification.
The command replaces the original \texttt{ros2 launch} call that is used to start the system.

In a federated deployment, \texttt{ros2 lf launch} additionally starts the LF Runtime Infrastructure (RTI) and the configured federates, either co-located on one machine or distributed across hosts according to the grouping selected in \secref{sec:semantics-federation}.
The RTI coordinates the logical clocks of the federates, so the determinism guarantee covers the entire deployment rather than a single process.
Our tool supports per-node federation (one federate per \ros{} node), and also any custom grouping of federates that are on the same system, as part of the compile step configuration.

\label{last-page}
\bibliographystyle{abbrv}
\bibliography{real-time}

\end{document}

%% file: figures/running_example_ros2.tex
\begin{tikzpicture}[
  node distance=1.0cm and 1.2cm,
  node box/.style={draw, rounded corners, align=center, font=\small,
                   inner sep=0.10cm},
  node A/.style={node box, minimum width=2.6cm, minimum height=1.5cm},
  node B/.style={node box, minimum width=1.4cm, minimum height=1.0cm},
  node D/.style={node box, minimum width=1.4cm, minimum height=1.7cm},
  cb/.style={draw, rounded corners, minimum width=0.7cm,
             minimum height=0.4cm, align=center, font=\footnotesize,
             fill=white, inner sep=1pt},
  timer/.style={draw, dashed, rounded corners, minimum width=0.8cm,
                minimum height=0.4cm, align=center, font=\footnotesize,
                fill=white, inner sep=1pt},
  outer/.style={draw, rounded corners, inner sep=0.35cm,
                line width=0.6pt, dashed, font=\footnotesize\itshape},
  arrow/.style={->, >=stealth, thick},
  topic/.style={font=\footnotesize\ttfamily}
]
% \draw[help lines, gray!30] (-1,-3) grid (10,3); % debug grid

% Node A: timer left, cA right. Vertically centered between B and C.
\node[node A] (a) {};
\node[font=\small, anchor=north] at (a.north) [yshift=-0.05cm] {Node A};
\node[timer] (a-tmr) at ([xshift=-0.6cm, yshift=-0.40cm]a.center) {timer $T$};
\node[cb]    (a-cA)  at ([xshift=0.55cm,  yshift=-0.40cm]a.center) {$c_A$};

% Node B: top middle, single callback cB.
\node[node B, above right=-0.5cm and 1.2cm of a] (b) {};
\node[font=\small, anchor=north] at (b.north) [yshift=-0.05cm] {Node B};
\node[cb] (b-cB) at ([yshift=-0.30cm]b.center) {$c_B$};

% Node C: bottom middle, single callback cC.
\node[node B, below right=-0.5cm and 1.2cm of a] (c) {};
\node[font=\small, anchor=north] at (c.north) [yshift=-0.05cm] {Node C};
\node[cb] (c-cC) at ([yshift=-0.30cm]c.center) {$c_C$};

% Node D: right, two callbacks stacked (the join node), centered between B and C.
\node[node D] (d) at ($(b.east)!0.5!(c.east) + (2.6cm,0)$) {};
\node[font=\small, anchor=north] at (d.north) [yshift=-0.05cm] {Node D};
\node[cb] (d-cD1) at ([yshift=0.20cm]d.center) {$c_{D1}$};
\node[cb] (d-cD2) at ([yshift=-0.40cm]d.center) {$c_{D2}$};

% Inside Node A: timer triggers cA (horizontal right).
\draw[arrow] (a-tmr.east) -- (a-cA.west);

% A.cA -> B.cB on alpha (upper path).
\draw[arrow] (a-cA.east) -- node[above, topic, pos=0.55, sloped] {/alpha} (b-cB.west);

% A.cA -> C.cC on beta (lower path).
\draw[arrow] (a-cA.east) -- node[below, topic, pos=0.55, sloped] {/beta} (c-cC.west);

% B.cB -> D.cD1 on gamma (upper path).
\draw[arrow] (b-cB.east) -- node[above, topic, pos=0.45, sloped] {/gamma} (d-cD1.west);

% C.cC -> D.cD2 on delta (lower path).
\draw[arrow] (c-cC.east) -- node[below, topic, pos=0.45, sloped] {/delta} (d-cD2.west);

% Outer system box wraps all four node boxes.
\node[outer, fit=(a) (b) (c) (d)] (sys) {};
\node[font=\footnotesize\itshape, anchor=north west]
      at ([xshift=0.12cm, yshift=-0.06cm]sys.north west) {running\_example};
\end{tikzpicture}

%% file: figures/pipeline.tex
\begin{tikzpicture}[
  pipeartifact/.style={draw, dashed, rounded corners, fill=gray!15,
                       minimum width=2.9cm, minimum height=0.65cm,
                       align=center, font=\small},
  pipestep/.style={draw, rounded corners, fill=blue!15,
                   minimum width=2.9cm, minimum height=0.75cm,
                   align=center, font=\small},
  arrow/.style={->, >=stealth, thick},
  label/.style={font=\footnotesize\itshape}
]
% Left column: top to bottom
\node[pipeartifact] (pkgs)                    {ROS~2 packages};
\node[pipestep,     below=0.4cm of pkgs]      (extract)  {Extraction\\\texttt{\footnotesize colcon lf-analyze}};
\node[pipeartifact, below=0.4cm of extract]   (json)     {\texttt{system.json}};
\node[pipestep,     below=0.4cm of json]      (gen)      {Generation\\\texttt{\footnotesize colcon lf-generate}};

% Right column: bottom to top, aligned with left column rows
\node[pipeartifact, right=1.2cm of gen]       (lf)       {\texttt{\{System\}.lf}};
\node[pipestep,     above=0.4cm of lf]        (compile)  {Compilation\\\texttt{\footnotesize lfc} / \texttt{\footnotesize colcon lf-build}};
\node[pipeartifact, above=0.4cm of compile]   (bin)      {C++ binary};
\node[pipestep,     above=0.4cm of bin]       (run)      {Execution\\\texttt{\footnotesize ros2 lf launch}};

% Left column flows down
\draw[arrow] (pkgs)    -- (extract);
\draw[arrow] (extract) -- (json);
\draw[arrow] (json)    -- (gen);
% Cross over to the right column
\draw[arrow] (gen)     -- (lf);
% Right column flows up
\draw[arrow] (lf)      -- (compile);
\draw[arrow] (compile) -- (bin);
\draw[arrow] (bin)     -- (run);
\end{tikzpicture}